%% file: main.tex
\documentclass[journal]{IEEEtran}
\usepackage{graphicx}
\usepackage{float}
\usepackage[utf8]{inputenc}
\usepackage{amsmath}
\usepackage[english]{babel}
\usepackage{amsfonts}
\usepackage{amssymb}
\usepackage[linesnumbered,ruled]{algorithm2e}
\usepackage{algpseudocode}
\usepackage{epsfig}

\input{packages}

\newtheorem{theorem}{Theorem}
\newtheorem{definition}[theorem]{Definition}
\DeclareMathOperator*{\argmin}{argmin}

\ifCLASSINFOpdf
\else
\fi
\usepackage[caption=false,font=footnotesize]{subfig}
\begin{document}
%
\title{Generative Adversarial Networks with Decoder-Encoder Output Noise}
%
%
%

\author{Guoqiang Zhong,~\IEEEmembership{Member,~IEEE},
        Wei Gao,
        Yongbin Liu,
        Youzhao Yang 
\thanks{This work was supported by the National Key R\&D Program of China under Grant 2016YFC1401004, the Science
and Technology Program of Qingdao under Grant No. 17-3-3-20-nsh, the CERNET Innovation Project under Grant
No. NGII20170416, and the Fundamental Research Funds for the Central Universities of China.}
\thanks{Y. Liu and G. Zhong are with the Department of Computer Science and Technology, Ocean University of China, Qingdao 266100, China. G. Zhong is the corresponding author of this paper. His e-mail address is gqzhong@ouc.edu.cn.}
\thanks{Y. Yang is with Fudan Univesity.}
\thanks{Y. Fu is with the Department of Electrical and Computer Engineering, College of Computer and Information Science, Northeastern University,
Boston, MA 02115 USA.}
}

\maketitle

\begin{abstract}
In recent years, research on image generation methods has been developing fast. The auto-encoding variational Bayes method (VAEs) was proposed in 2013, which uses variational inference to learn a latent space from the image database and then generates images using the decoder. The generative adversarial networks (GANs) came out as a promising framework, which uses adversarial training to improve the generative ability of the generator. However, the generated pictures by GANs are generally blurry. The deep convolutional generative adversarial networks (DCGANs) were then proposed to leverage the quality of generated images. Since the input noise vectors are randomly sampled from a Gaussian distribution, the generator has to map from a whole normal distribution to the images. This makes DCGANs unable to reflect the inherent structure of the training data. In this paper, we propose a novel deep model, called generative adversarial networks with decoder-encoder output noise (DE-GANs), which takes advantage of both the adversarial training and the variational Bayesain inference to improve the performance of image generation. DE-GANs use a pre-trained decoder-encoder architecture to map the random Gaussian noise vectors to informative ones and pass them to the generator of the adversarial networks. Since the decoder-encoder architecture is trained by the same images as the generators, the output vectors could carry the intrinsic distribution information of the original images. Moreover, the loss function of DE-GANs is different from GANs and DCGANs. A hidden-space loss function is added to the adversarial loss function to enhance the robustness of the model. Extensive empirical results show that DE-GANs can accelerate the convergence of the adversarial training process and improve the quality of the generated images.
\end{abstract}


%
\IEEEpeerreviewmaketitle

\section{Introduction}
\label{1}
%
%
%
%
\IEEEPARstart{I}{mage} generation is an interesting, important but intractable task. Before deep learning approaches are applied in this area, traditional methods based on probability theory and Shannon's information theory have been explored for a long time. The non-parametric sampling algorithm was presented and applied in texture synthesis~\cite{efros1999texture}. It assumes a Markov random field and estimates new pixels using the conditional distribution of their neighbors. In~\cite{portilla2000parametric}, Portilla and Simoncelli proposed a parametric texture model based on joint statistics, which uses a decomposition method that is called steerable pyramid decomposition to decompose the texture of images. An example-based super-resolution algorithm~\cite{freeman2002example} was proposed in 2002, which uses a Markov network to model the spatial relationship between the pixels of an image. A scene completion algorithm~\cite{hays2007scene} was proposed in 2007, which applied a semantic scene match technique. These traditional algorithms can be applied to particular image generation tasks, such as texture synthesis and super-resolution. Their common characteristic is that they predict the images pixel by pixel rather than generate an image as a whole, and the basic idea of them is to make an interpolation according to the existing part of the images. Here, the problem is, given a set of images, can we generate totally new images with the same distribution of the given ones?

\indent Thanks to the development of deep learning, we already have the ability to extract features from images using deep neural networks~\cite{dosovitskiy2017learning},~\cite{huang2017beyond},~\cite{gregor2015draw},~\cite{wei2002multisynapse},~\cite{setiono2002extraction},~\cite{xing2017deep},~\cite{goh2014learning},~\cite{du2017learning}. One of the common models is the stacked auto-encoder~\cite{masci2011stacked} which is used in unsupervised learning. The training images are firstly passed into the encoder and transformed into the latent space. Then the decoder tries to reconstruct the images from the compact latent space. The training goal of the auto-encoder is to minimize the reconstruction-error between the reconstructed images and the original images. In 2013, the auto-encoding variational Bayes method (VAEs)~\cite{kingma2013auto} came out as a promising generative model. It shares a similar structure as the ordinary auto-encoders but a significant different point is that the encoder firstly maps the images to a Gaussian distribution and the decoder could map from Gaussian noise to the generated images. Hence, a trained decoder of VAEs could directly transform a Gaussian noise vector to an image. Another novel deep learning attempt, which was built on the nonequilibrium thermodynamics, was presented in~\cite{sohl2015deep}. However, an obvious defect of these models is that the images generated by them are blurry and the edges are difficult to distinguish.

\indent Some pieces of work, such as the image generation with PixelCNN-Decoders~\cite{van2016conditional}, are proposed to combine the local details and the holistic shapes to improve the quality of generated images. For concreteness, PixelCNN-Decoders assume an conditional distribution of the neighborhood pixels and adjusts the architecture of PixelCNNs. However, these approaches can only generate images pixel by pixel, but not generate the images as a whole.

\indent In 2014, a pioneering work was proposed which applied an adversarial strategy and is known as Generative Adversarial Networks(GANs)~\cite{goodfellow2014generative}. The GANs framework is composed of a generator \textit{G} and a discriminator \textit{D}. The generator tries to confuse the discrimintor by generating images as plausible as possible, and the discriminator tries to distinguish the fake generated images from the real ones. They are trained alternatively to achieve the convergence. One of the notable distinctions between GANs and the traditional generative models is that the GANs generate an image as a whole rather than pixel by pixel. In the original GANs framework, a generator is made up of fully connected neural networks including two dense layers and a dropout layer. Random noise vectors are sampled from a normal distribution and delivered as inputs to the generator networks. The discriminator can be any supervised learning model. Image generation techniques based on GANs explode recently, such as the generative model using a Laplacian Pyramid of adversarial networks~\cite{denton2015deep}. It generates images with relatively high resolution in spite of its unstable generation process. The idea of adversarial training has also been widely applied. For instance, the adversarial autoencoders~\cite{makhzani2015adversarial}, adversarial variational Bayes~\cite{mescheder2017adversarial} and the work of~\cite{larsen2015autoencoding} are designed to generate images based on the adversarial training, with meritorious results. PixelGAN autoencoder~\cite{makhzani2017pixelgan} is another noticeable work that makes an attempt to explore the input noise vectors. The generative path of PixelGAN autoencoder is a PixelCNN~\cite{van2016conditional} that is conditioned on a latent code, while its recognition path uses a GAN to impose a prior distribution on the latent code.

\indent In order to accelerate the training of GANs and explore the reason why the training process is not stable, Wasserstein GAN (WGAN)~\cite{arjovsky2017wasserstein} was proposed. It proves that the defect of the loss function of GANs is that it cannot ensure the convergence of training. Hence, in WGAN, the Wasserstein distance was used to replace the original loss function of GANs. However, the calculation of Wasserstein distance is complex and slow. Inspired by the WGAN model, we also enhance our loss function to improve the robustness of the model and accelerate the convergence, but we do not adopt the Wasserstein distance.

\indent Diversity is an important criterion on the generated images. DeLiGAN~\cite{gurumurthy2017deligan} explores image generation on a limited dataset and assumes the latent space to be a mixture of Gaussian distributions. This assumption could relatively improve the diversity of the generated images. Moreover, \cite{gurumurthy2017deligan} proposed the concept of modified-inception-score to measure the diversity of the generated images.

\indent Based on the original GANs, the study of the structure of the generators makes great process. Deconvolutional networks were used as the generator~\cite{dosovitskiy2015learning,radford2015unsupervised} and the performance turned out to be favorable. The deep convolutional generative adversarial networks (DCGANs)~\cite{radford2015unsupervised} uses deep CNNs~\cite{krizhevsky2012imagenet} as the discriminator and deconvolutional neural networks~\cite{zeiler2010deconvolutional} as generator. 100-dimensional noise vectors were sampled from a normal distribution and then transmitted to the deconvolutional networks as inputs. However, the convergence of DCGANs is not fast and the quality of generated images is not good enough.

\indent In both GANs and DCGANs models, the noise vectors are simply sampled from a normal distribution \textit{N}(0, $\mathbf{I}$), which may cause a deviation from the real distribution of real images. Hence, in this paper we are to find a method to learn a better distribution of noise. Inspired by the structure of VAEs, we present a novel Decoder-Encoder architecture to fit a noise distribution which carries the information of the image datasets.

\indent The Decoder-Encoder structure is adjusted from the VAEs networks. We first trained a VAEs model using an image dataset to make the networks carry information of the image datasets, and then swap the trained decoder and encoder. A Gaussian noise vector is sent to the decoder and flows through the whole architecture. Since the dimension of the input of the decoder and the output of the encoder are the same, the Decoder-Encoder structure doesn't change the dimension of the noise vectors but makes them carry the information of the images. It transforms a Gaussian noise to a posterior one which is used as the input of the DCGANs generator. The experiment demonstrates that the posterior noise is better since it can accelerate the convergence of the adversarial networks and improve the quality of generated images.

\indent Besides the study of noise, we also enhance the loss function by adding a hidden-space loss functions to the original adversarial loss in the traditional GANs model. Experiments indicate an obvious increase on quality of the generated images using the combined loss function.

\indent Our contributions are as follows:
\begin{itemize}
    \item We propose a Decoder-Encoder structure which transforms the original Gaussian noise to a posterior one and it can carry information of the real images. These noise vectors could preserve the inherent structure of images, accelerate the training process and improve the quality of the generated images.
    \item We add a hidden-space loss function to the adversarial loss function of the traditional GANs and DCGANs, which can improve the quality of the generated images.
    \item We propose the generative adversarial networks with decoder-encoder structure (DE-GANs) model. It generates images with higher quality and algorithm efficiency compared to the GANs and DCGANs model.
\end{itemize}
\indent The rest of this paper is organized as follows. In Section \ref{2}, we introduce the proposed DE-GANs model, including the Decoder-Encoder architecture and the training algorithm of DE-GANs. In Section \ref{3}, we analyze the distribution and dimension of the noise used in DE-GANs. Experiments are reported in Section \ref{4} and conclusion is drawn in Section \ref{5}.

\section{Generative Adversarial Networks with Decoder-Encoder Output Noise}
\label{2}
In this section, we present the proposed DE-GANs model. Firstly, we introduce the Decoder-Encoder architecture, which generate informative noise to the generator of DE-GANs, and then we specify the network structure of DE-GANs and their learning algorithm.

\subsection{Decoder-Encoder Architecture}
Auto-encoders (AEs)~\cite{bengio2013representation} are common deep models in unsupervised learning. It aims to represent a set of data by the latent layers. Architecturally, AEs consist of two parts, the encoder and decoder. The encoder part takes the input $\x\in \mathbb{R}^d$ and maps it to $\z$ which is called latent variable. And the decoder tries to reconstruct the input data from $\z$. The training process of autoencoders is to minimise the reconstruction error. Formally, we can define the encoder and the decoder as transitions $\tau_1$ and $\tau_2$:

\begin{align}
\label{eq:1}
\tau_1(X) &\rightarrow Z \\
\tau_2(Z) &\rightarrow \hat{X}\\
\tau_1,\tau_2  &= \argmin_{\tau_1,\tau_2} {\Vert X-\hat{X} \Vert}^2
\end{align}

The VAEs model shares the same structure with the autoencoders, but it's based on an assumption that the latent variables follow some kind of distribution, such as Gaussian or uniform distribution. It uses variational inference for the learning of the latent variables. In VAEs the hypothesis is that the data is generated by a directed graphical model $p(\x|\z)$ and the encoder is to learn an approximation $q_{\phi}(\z|\x)$ to the posterior distribution $p_{\theta}(\z|\x)$. The training process of VAEs is to minimize the loss function:
\begin{equation}
\label{eq:2}
\mathfrak{L} = D_{KL}(q_\phi(\z|\x)||p_\theta(\z))-\mathbb{E}_{q_\phi}(\log p_\theta(\x|\z)).
\end{equation}
In equation \ref{eq:2} $D_{KL}$ stands for the Kullback-Leibler divergence~\cite{kullback1951information}.

\indent In view of the mechanism of VAEs, we can transform stochastic vectors $\z$ sampled from a normal distribution $N$(0, \I) to posterior ones using a VAEs decoder trained by image datasets. According to the assumption of VAEs, the posterior vectors meet an approximate distribution of real images and if we connect posterior vectors to the encoder which is part of the same VAEs as decoder, we can map the posterior vectors to new noise vectors $\z^\prime$, which carry information of real images. The vectors sampled from a bare Gaussian distribution include a number of noise points which are not suitable for image generation. So, if we use noise $\z^\prime$ as the input of the generator of DCGANs, we can eliminate the unreasonable noise points and improve the quality of generated images. Mathematically, the process can be described as follows:

VAEs training stage
\begin{align}
\label{eq:3}
\theta:& \x\rightarrow \z, \x \sim p(\x), \z \sim p(\z|\x) \\
\label{eq:4}
\phi:& \z \rightarrow \hat{\x}, \z \sim p(\z|\x) , \hat{\x} \sim p(\x|\z) \\
\label{eq:5}
\theta,\phi =& \argmin_{\theta,\phi} D_{KL}(q_\phi(\z|\x)||p_\theta(\z))-\mathbb{E}_{q_\phi}(\log p_\theta(\x|\z))
\end{align}
In expressions \ref{eq:3} and \ref{eq:4}, $\theta$ is the transition of encoders and $\phi$ denotes the mapping from latent variable distribution $p(\z|\x)$ to approximate distribution $p(\x|\z)$.

Decoder-Encoder stage
\begin{align}
\label{eq:6}
\phi:& \z_n \rightarrow \hat{\x}, \z_n \sim \emph{N}(0, \I), \hat{\x} \sim p(\x|\z) \\
\label{eq:7}
\theta:& \hat{\x} \rightarrow \z^\prime, \z^\prime \sim p(\z|\x)
\end{align}
In expressions \ref{eq:6} and \ref{eq:7}, $\phi$, $\theta$ are identical to equation \ref{eq:5}. As presented above, the Decoder-Encoder structure integrates the information of images into Gaussian vectors to make the noise vectors more suitable to generate images by DCGANs.

\begin{figure}
\centering
\includegraphics[width=\linewidth,height=\textheight,keepaspectratio]{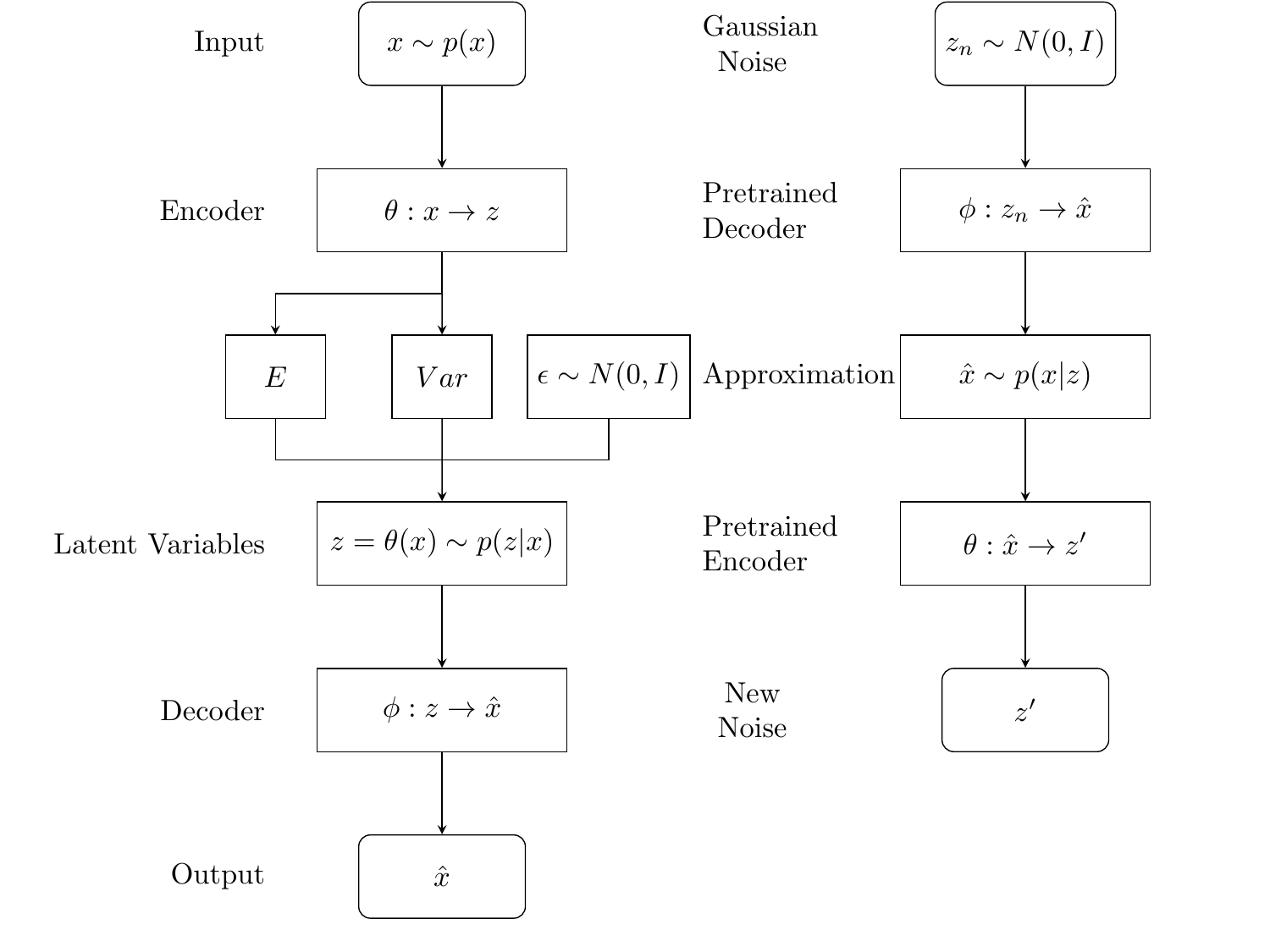}
\caption{The left is standard VAEs model and the right is our Decoder-Encoder structure. The decoder and encoder pre-trained in VAEs are swapped. Noise vectors sampled from Gaussian distribution are sent into the decoder and eventually transformed to vectors which carry information of image data.}
\end{figure}
\begin{figure*}[t]
\centering
\includegraphics[page=1,width=\linewidth,height=\textheight,keepaspectratio]{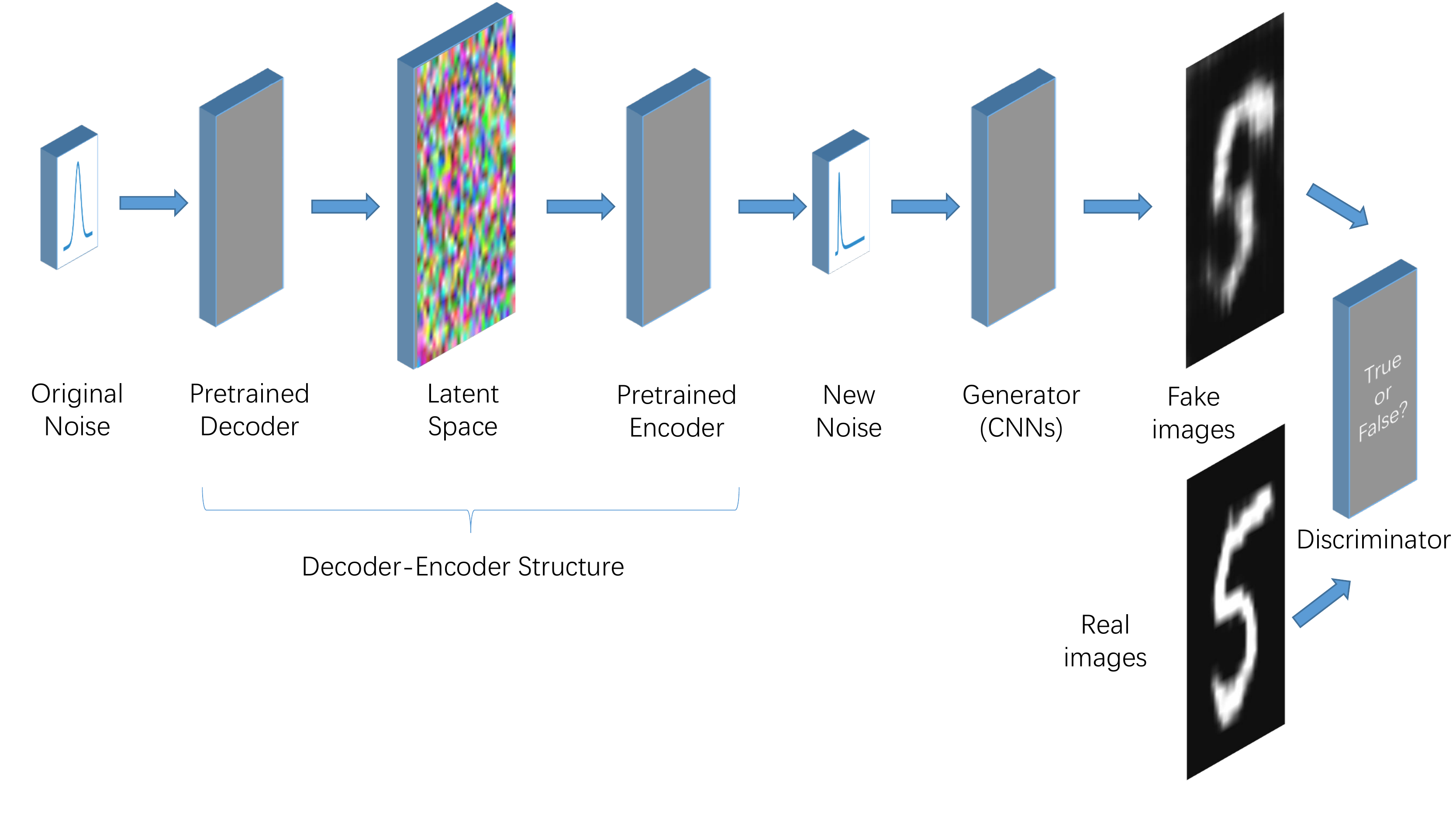}
\caption{The DE-GANs architecture. Its generator and discriminator are composed of convolutional neural networks whereas a pre-trained Decoder-Encoder is combined with the generator to add prior information to the input noise vectors.}
\label{figure:GAN}
\end{figure*}

\subsection{DE-GANs Architecture}
Our DE-GANs architecture comprises two parts, the preconditioner (pre-trained Decoder-Encoder structure) and the DCGANs component. The new noise vectors containing information of real images are delivered into the generator of DCGANs to generate fake images and the discriminator tries to distinguish the fake images from real images. The stochastic gradient descent algorithm (SGD)~\cite{dorigo2006ant} is adopted to optimize the model. Furthermore, we apply the batch-normalization (BN)~\cite{ioffe2015batch} to the output of each convolutional layer to ensure all the output variables follow a Gaussian distribution. The BN layers are effective to reduce coupling of adjacent layers and accelerate the training process. Deconvolutioal networks are used as the whole generator which maps the input noise vectors to the output ``fake" images.

\indent The discriminator uses deep convolutional networks to distinguish the fake images from real images. Real images sampled from datasets are labeled as 1 while the fake as 0 and these images are sent to the discriminator as inputs. Each output corresponding to an input figure is a sigle scalar $D(\x)$ which denotes the probability that the input image $\x$ is a real image. The loss function of a discriminator is as follows:

\begin{equation}
\label{eq:10}
\begin{split}
 \mathfrak{L}_D &= -log(|L_\x-D(\x)|) \\
  & = -(log(1-D(\x_{real})) + log(D(G(\z_n)))).
\end{split}
\end{equation}
In equation \ref{eq:10}, $L_\x$ denotes the label of image $\x$, and $\x_{real}$ represents a real image.

\subsection{Loss Function}
In order to improve the quality of generated images, we modify the loss function of DCGANs to a combination of three loss functions including the adversarial loss function, the squared loss function and the hidden-space loss function. The adversarial loss function is:
\begin{equation}
\begin{split}
    \mathfrak{L}_{adv} &= -\mathfrak{L}_D\\
    &= \mathbb{E}_{\x\sim p_{data}(\x)}(\log D(\x)) + \mathbb{E}_{\z\sim p_\z(\z)}(\log (1-D(G(\z)))).
\end{split}
\end{equation}
The adversarial loss function $L_{adv}$ reflects the idea of competition since we have to simultaneously reinforce the discriminator and the generator in order to minimize the $L_{adv}$.

\indent Besides the adversarial loss function, the hidden-space loss function is proposed to enhance the training algorithm. In order to improve the robustness of the algorithm, we use the hidden-space loss function as a metric of the discrepancy of images. The hidden-space loss function takes activation maps of latent variables to compute the divergence of images on the basis of the fact that the latent layers extract features of the input images which is invariant to rotation and translation.

\begin{equation}
\label{eq:13}
\mathfrak{L}_{hid}  = \frac{1}{N}\sum_{i = 1}^{N}\Vert h^{(i)}(\x_{real})-h^{(i)}(\x_{gen}) \Vert.
\end{equation}

\begin{figure*}[t]
\centering
\includegraphics[width=\linewidth,height=\textheight,keepaspectratio]{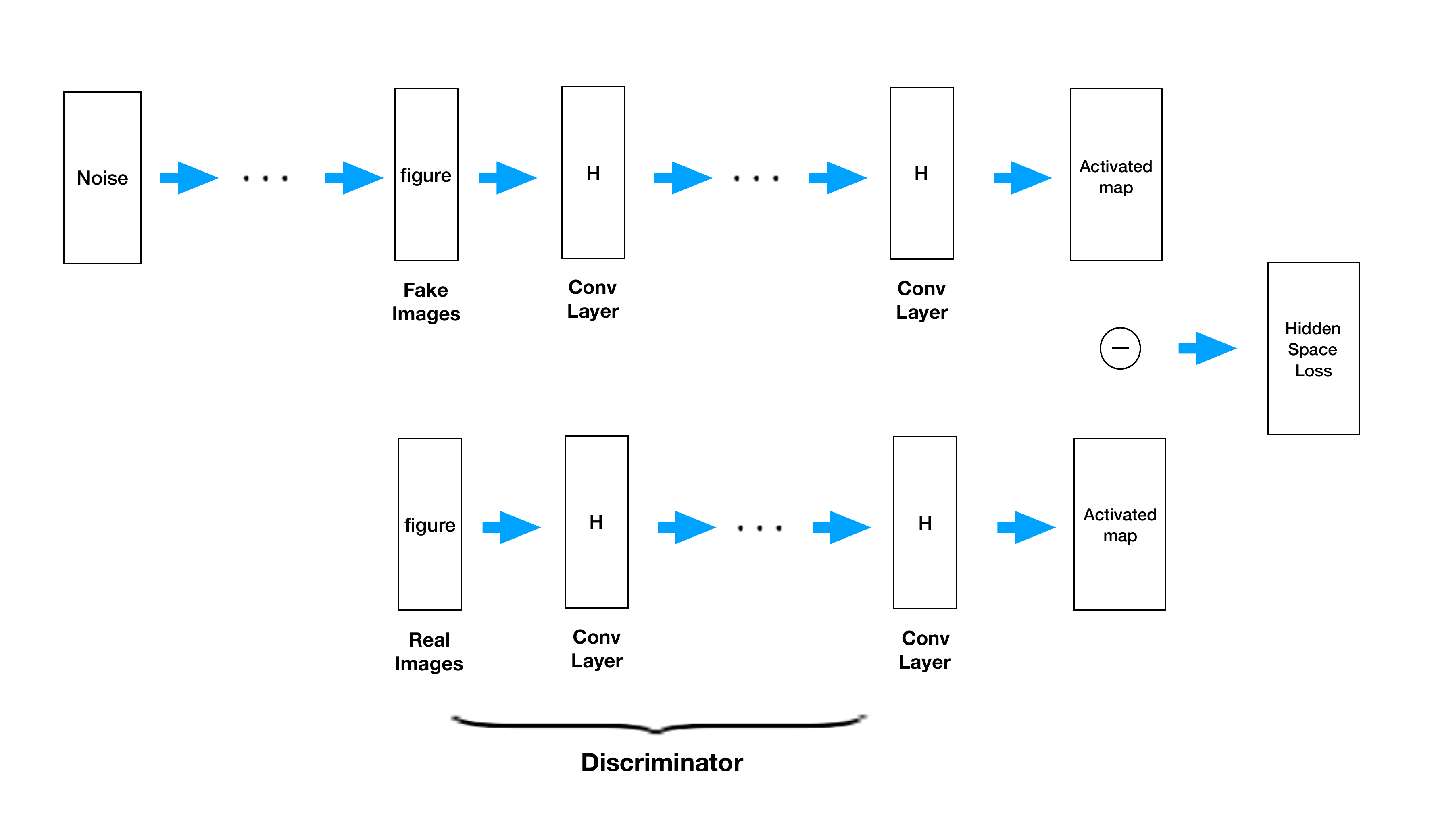}
\caption{When a real image is fed into the discriminative convolutional networks, $h(\x_{real})$ is the output as opposed to a generated image with $h(\x_{gen})$ as output.}
\label{fig:H_loss}
\end{figure*}

In equation \ref{eq:13}, $h(\x_{real})$ represents the activation map~\cite{zeiler2014visualizing} of a high convolutional layer when a real image is sent to the discriminator, while $h(\x_{gen})$ denotes a generated(fake) one. A clear computation is demonstrated in figure \ref{fig:H_loss}.

\indent Eventually the combined loss function is
\begin{equation}
\mathfrak{L} = \lambda_1 \mathfrak{L}_{adv} + \lambda_2 \mathfrak{L}_{hid},
\end{equation}
where $\lambda_1, \lambda_2$ denote the weights of the three loss functions.

Specifically, the training procedure of DE-GANs is shown in Algorithm~\ref{al:1}.
\begin{algorithm}\label{al:1}
    \SetKwInOut{Input}{Input}
    \SetKwInOut{Output}{Output}

    \Input{$\alpha$: learning rate, $N$: batch size, $\lambda_1, \lambda_2$: weights of $\mathfrak{L}_{adv}, \mathfrak{L}_{hid}$, $X$ real image datasets.}

    \Output{$\Gamma$: Latent variables of VAEs, $\theta$: latent variables of the discriminator, $\phi$: latent variables of the generator.}

    Pre-train VAEs using $X$ and determine $\Gamma$ \;
    \Repeat{deadline}{
      Sample $\x_{real}$ from dataset \;
      Sample $\z_n$ from Gaussian distribution $N(0, \I)$\;
      $\hat{\x}$ $\leftarrow$ VAEs-Decoder($\z_n$)\;
      $\z^\prime$ $\leftarrow$ VAEs-Encoder($\hat{\x}$)\;
      $\x_{gen}$ $\leftarrow G(\z^\prime)$\;
      $h_{real} \leftarrow D(\x_{real})$ \;
      $h_{gen} \leftarrow D(\x_{gen})$ \;
      $\mathfrak{L}_{hid}  \leftarrow \frac{1}{N}\sum_{i=1}^{N}\Vert h^{(i)}(\x_{real})-h^{(i)}(\x_{gen})\Vert $\;
      $\mathfrak{L}_{adv} \leftarrow \mathbb{E}_{\x\sim p_{data}(\x)}[\log D(\x)] + \mathbb{E}_{\z\sim p_\z(\z)}[\log (1-D(G(\z)))]$\;

      $\theta \leftarrow \theta + \alpha \nabla_\theta(\lambda_1 \mathfrak{L}_{adv}+\lambda_2 \mathfrak{L}_{hid})$

      $\phi \leftarrow \phi + \alpha \nabla_\phi(\lambda_1 \mathfrak{L}_{adv}+\lambda_2 \mathfrak{L}_{hid})$
    }

    \caption{Training Algorithm for DE-GANs}
\end{algorithm}

\section{Noise}
\label{3}
\begin{figure}[H]
\subfloat[]{
\includegraphics[width=\linewidth,height=\textheight,keepaspectratio]{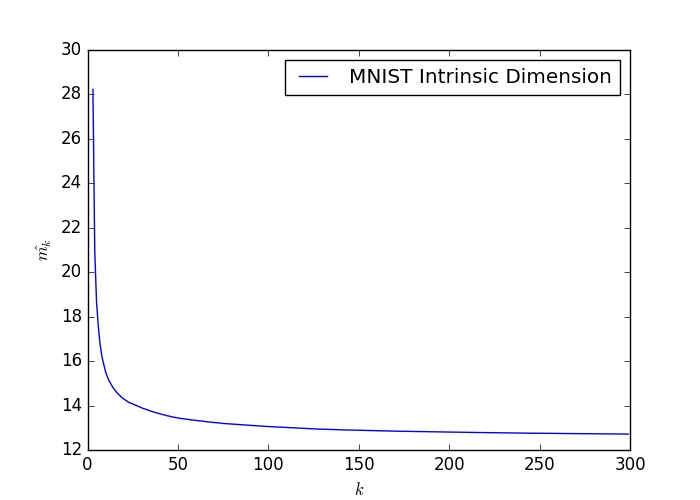}
\label{mnist_intrinsic_d}
}
\quad
\subfloat[]{\includegraphics[width=\linewidth,height=\textheight,keepaspectratio]{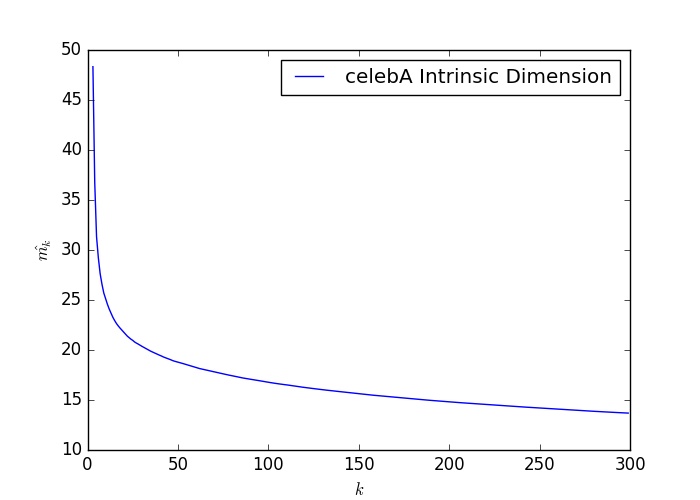}
\label{celebA_intrinsic_d}
}
\quad
\subfloat[]{\includegraphics[width=\linewidth,height=\textheight,keepaspectratio]{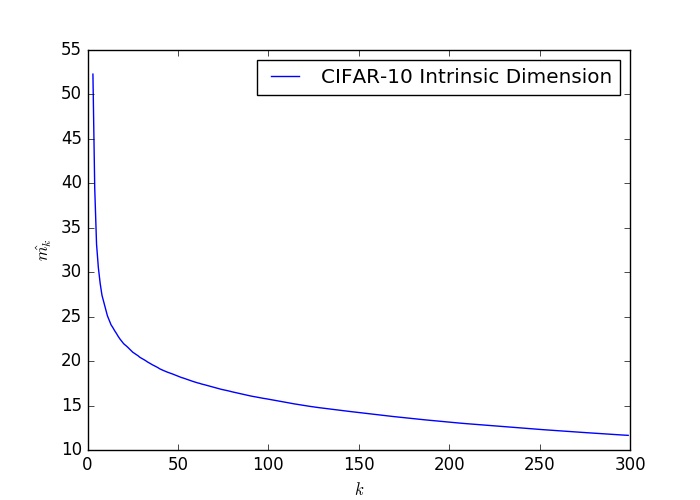}
\label{cifar_intrinsic_d}
}
\caption{This figure is a result of the maximum likelihood method to estimate the intrinsic dimension of the MNIST, CIFAR-10 and celebA data sets. The horizontal axis indicates the number of neighbor points that are used to estimate the intrinsic dimension and the vertical axis denotes the approximation of the intrinsic dimension. As the curves shows, when $k$ converges towards to infinity, the estimated intrinsic dimension of the dataset manifold will approach to a constant. This constant is the estimated intrinsic dimension (EID). As shown in subfigure \ref{mnist_intrinsic_d}, \ref{cifar_intrinsic_d}, and \ref{celebA_intrinsic_d}, the EID of MNIST and CIFAR-10 datasets are less than 13 and the celebA is less than 15. A more accurate value is presented in table \ref{intrinsic_accurate_values}.}
\label{fig:infrinsic_d}
\end{figure}

\indent In this section, we will elaborate our study on the dimension of input noise of the adversarial networks. The images can be reshaped to vectors which distribute over a manifold. And what a generator do is to map a noise vector to an image. Generally, the dimension of noise vectors are much smaller than that of images. How should we choose a suitable dimension for the noise vectors so that it will not cause information loss during the mapping?
\indent The least number of variables to represent (which includes linear and nonlinear mapping) all the elements in a specific vector space is called the \emph{intrinsic dimension} of it~\cite{bennett1969intrinsic}. Mathematically, it can be defined as follows~\cite{yu2009nonlinear}:
\begin{definition}
A subset $T$ of $\mathbb{R}^n$ is called a $P$-smooth manifold with intrinsic dimension $k$ if there exists a constant $C_p$ such that for any $x \in T$ there exists a set of vectors $v_1(\x),\cdots, v_k(\x) \in \mathbb{R}^n$ s.t.
\begin{equation}
\inf_{\gamma \in \mathbb{R}^k}{\Vert x^\prime -x -\sum_{j=1}^{k}\gamma_jv_j{x}\Vert} \leq C_p\Vert x^\prime -x \Vert
\end{equation}
is true for any $\x^\prime \in T$.
\end{definition}
\indent If the dimension of noise vectors is lower than the intrinsic dimension of images then the generator will never map noise vectors to the whole distribution of real data. It causes loss of information which is called \emph{mode collapse} and the generated images will seem to be homoplasy. In order to avoid the mode collapse, estimation of the intrinsic dimension of datasets is necessary. In this paper we use a maximum likelihood method to solve this task~\cite{levina2005maximum}:
\begin{definition}
If $T$ is a manifold and $\{x_i\}_{i=1}^N$ denotes $N$ points sampled from $T$, then
\begin{equation}
\hat{m}_k(x_i)=(\frac{1}{k-1}\sum_{j=1}^{k-1}\log \frac{T_k(\x_i)}{T_j(\x_i)})^{-1},
\end{equation}
\begin{equation}
\hat{M}_k = \frac{1}{N}\sum_{i=1}^N\hat{m}_k(\x_i),
\end{equation}
\begin{equation}
\hat{M} = \lim_{k \to \infty}\hat{M}_k,
\end{equation}
where $T_k(x_i)$ denotes the distance(vector norm) between the $k$-th neighbor and $\x_i$ and $\hat{M}_k$ denotes the estimated intrinsic dimension when we only consider the nearest $k$ neighbors of a certain point.
\end{definition}
\indent We apply this algorithm to three image datasets, \emph{MNIST}~\cite{MNIST},\emph{celebA}~\cite{liu2015faceattributes}  and \emph{CIFAR-10}~\cite{krizhevsky2009learning} and as the result shown in figure \ref{fig:infrinsic_d}, the estimation of intrinsic dimension of MNIST, celebA and CIFAR-10 are about 13, 15 and 13, respectively. The accurate values are presented in table \ref{intrinsic_accurate_values}. If the dimension of noise vectors is higher than the intrinsic dimension, there shouldn't exist mode-collapsing problem and the 100-dimensional noise vectors can fit the condition.

\indent On the other hand, if the noise dimension is too large, it will introduce too much noise and affect the quality of the generated images. In our experiments, we set the dimension of noise vectors to 128.

\begin{table}[!t]
\begin{center}
\begin{tabular}{ c c c }
MNIST & CIFAR-10 & celebA\\
\hline
 12.7 & 11.6 & 13.7
\end{tabular}
\end{center}
\caption{The accurate values of the estimated intrinsic dimensionalities of the \emph{MNIST, CIFAR-10} and \emph{celebA} data sets.}
\label{intrinsic_accurate_values}
\end{table}

\section{Experiments}
\label{4}
In order to demonstrate the effectiveness of our DE-GANs architecture, we apply the model to three datasets MNIST, celebA and CIFAR-10. As discussed in section \ref{3}, the dimension of the noise vectors is determined to be 128 in order to avoid mode collapse. The sampled noise vectors meet Gaussian distribution with zero mean and uniform variance. Considering the intrinsic structures of the three datasets are different, we adapt the specific architectures respectively. The following discussion will mainly be divided into three parts according to the categories of datasets.

\begin{itemize}
\item In Subsection \ref{4.1}, we use the MNIST dataset as the real images and compare its convergence rate with other generative models. The results show that our DE-GANs model could approach the convergence faster than the state-of-art models. The specific architecture of the adversarial networks is shown in table \ref{MNIST_archi}
\item In Subsection \ref{4.2}, we apply the model to the celebA dataset and get a series of changing facial images which show how the images change with the noise vectors intuitively. Since the celebA images are in size of $218\times 178\times 3$, which isn't tractable for training, we cut and resize the images to size of $64\times 64\times 3$. The structure of the adversarial networks is shown in table \ref{celebA_archi}
\item In Subsection \ref{4.3}, we apply the adversarial networks to CIFAR-10 dataset and generate some tiny images. The specific structure of the adversarial networks is shown in table \ref{CIFAR_archi}
\end{itemize}
\begin{table*}[t]
    \centering
    \begin{tabular}{c c c c}
      Decoder &  Encoder & Generator & Discriminator\\
     \hline
      $Batch Size\times 128$ & $Batch Size\times 28\times 28 $ & $Batch Size\times 128$ & $Batch Size \times 28 \times 28$ \\
      $3\times 3\times 128 \uparrow$ & $5\times 5\times 32 \downarrow$ & $1024$ Fully-Connected+BatchNorm+relu & $5\times 5\times 11\downarrow$ lrelu \\
      $5\times 5\times 64 \uparrow$ & $5\times 5\times 64 \downarrow$ & $6272$ Fully-Connected+BatchNorm+relu  & $5\times 5\times 74\downarrow$ BatchNorm+lrelu \\
      $5\times 5\times 32\uparrow$ & $5\times 5\times 128\downarrow$ & $5\times 5\times 128\uparrow$ BatchNorm+relu & 1024 Fully-Connected+BatchNorm+lrelu \\
      $5\times 5 \uparrow$ & Dropout & $1\times 5\times 5 \uparrow$ & $1$ Fully-Connected\\
      Sigmoid & Flatten & Sigmoid & Sigmoid \\
      Flatten & $128$ Fully-Connected & $BatchSize\times 28\times 28$ Output & $BatchSize\times 1$ Output \\
      $BatchSize\times 784$ Output & $BatchSize\times 128$ Output & &\\
            \hline
    \end{tabular}
    \caption{DE-GANs architectures applied to MNIST (a $\uparrow$ means an upsampler using convolution operation while a $\downarrow$ denotes a downsampler)}
    \label{MNIST_archi}
\end{table*}
\begin{table*}[t]
    \centering
    \begin{tabular}{c c c c}
      Decoder &  Encoder & Generator & Discriminator\\
     \hline
      $BatchSize\times 128$ & $BatchSize\times 64\times 64 \times 3 $ & $BatchSize\times 128$ & $BatchSize\times 64 \times 64 \times 3$ \\
      $4\times 4\times 128 \uparrow$ & $5\times 5\times 32 \downarrow$ & $1024 $ Fully-Connected+BatchNorm+relu & $5\times 5\times 64\downarrow$ lrelu \\
      $5\times 5\times 64 \uparrow$ & $5\times 5\times 64 \downarrow$ & $5\times 5\times 512 \uparrow$ BatchNorm+relu  & $5\times 5\times 128\downarrow$ BatchNorm+lrelu \\
      $5\times 5\times 32\uparrow$ & $5\times 5\times 128\downarrow$ & $5\times 5\times 256\uparrow$ BatchNorm+relu & $5\times 5\times 256\downarrow$ BatchNorm+lrelu \\
      $5\times 5 \times 16\uparrow$ & Dropout & $5\times 5 \times 512 \uparrow$ & 1 Fully-Connected\\
      $5\times 5\times 3\uparrow$+tanh & Flatten & $5\times 5\times 64\uparrow$ & 64 Fully-Connected \\
      Flatten & $128$ Fully-Connected & tanh & Sigmoid \\
      $BatchSize\times12288$ Output & $BatchSize\times 128$ Output & $BatchSize\times 64\times 64 \times 3$ Output & $BatchSize \times 1 $ Output\\
      \hline
    \end{tabular}
    \caption{DE-GANs architectures applied to celebA (a $\uparrow$ means an upsampler using convolution operation while a $\downarrow$ denotes a downsampler)}
    \label{celebA_archi}
\end{table*}
\begin{table*}[t]
    \centering
    \begin{tabular}{c c c c}
      Decoder &  Encoder & Generator & Discriminator\\
     \hline
      $BatchSize\times 128$ & $BatchSize\times 32\times 32 \times 3 $ & $BatchSize\times 128$ & $BatchSize\times 32 \times 32 \times 3$ \\
      $4\times 4\times 128 \uparrow$ & $5\times 5\times 32 \downarrow$ & $1024 $ Fully-Connected+BatchNorm+relu & $5\times 5\times 64\downarrow$ lrelu \\
      $5\times 5\times 64 \uparrow$ & $5\times 5\times 64 \downarrow$ & $5\times 5\times 512 \uparrow$ BatchNorm+relu  & $5\times 5\times 128\downarrow$ BatchNorm+lrelu \\
      $5\times 5\times 32\uparrow$ & $5\times 5\times 128\downarrow$ & $5\times 5\times 256\uparrow$ BatchNorm+relu & $5\times 5\times 256\downarrow$ BatchNorm+lrelu \\
      $5\times 5 \times 3\uparrow$ & Dropout & $5\times 5 \times 512 \uparrow$ & 1 Fully-Connected\\
      tanh & Flatten & $5\times 5\times 64\uparrow$ & 64 Fully-Connected \\
      Flatten & $128$ Fully-Connected & tanh & Sigmoid \\
      $BatchSize\times3072$ Output & $BatchSize\times 128$ Output & $BatchSize\times 32\times 32 \times 3$ Output & $BatchSize \times 1 $ Output\\
      \hline
    \end{tabular}
    \caption{DE-GANs architectures applied to CIFAR-10 (a $\uparrow$ means an upsampler using convolution operation while a $\downarrow$ denotes a downsampler)}
    \label{CIFAR_archi}
\end{table*}

\subsection{Experiments on MNIST Dataset}

Experiments in this subsection mainly present the generated images and the evaluation of the quality of these images as well as the convergence rate of the training process.
\label{4.1}
\begin{figure}
\includegraphics[width=\linewidth,height=\textheight,keepaspectratio]{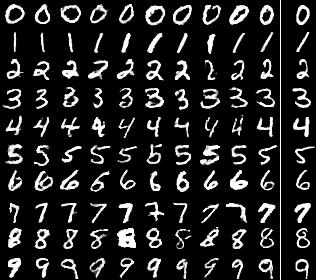}
\caption{Samples from the generated images of MNIST digits. The digits in the last column is the closest ones in original MNIST dataset to the last but one column (measured by Euclidean distance).}
\end{figure}

\subsubsection{How we measure the quality of the generated digit images}
In order to estimate the quality of the generated images of DE-GANs and compare it with other generative models. We use a strategy to evaluate the quality, that is whether human's eyes can recognize the generated digits. However, people's recognition may be not objective and may be influenced by their cognitive ability or light intensity. Hence, we firstly trained a supervised CNN classifier using the MNIST handwritten digit database. The test classification error is less than 0.005 so it can substitute human's eyes to evaluate whether a handwritten digit can be recognized. We randomly take a set of samples which contain 6400 generated digit images of our DE-GANs model and the test accuracy is $99.1\%$. For comparison, we train some other state-of-art generative models using the MNIST handwritten digits and the test accuracy results are presented in table \ref{table:comparison}

\begin{table}[t]
\begin{center}
\begin{tabular}{ c | c }
\hline
Model & Test Accuracy\\
\hline
DCGAN & $99.0\%$ \\
WGAN~\cite{arjovsky2017wasserstein} & $94.3\%$ \\
VAE & $97.7\%$\\
GAN & $69.7\%$\\
AAE~\cite{makhzani2015adversarial} & $97.8\%$\\
\hline
DE-GANs & $99.1\%$\\
\hline
\end{tabular}
\end{center}
\caption{The parameters of all the models are optimized to obtain the best performance. The test accuracy reflects the quality of the generated MNIST handwritten digits. A higher test accuracy means the generated digit images are much easier to recognized that is to say the model generates higher quality images.}
\label{table:comparison}
\end{table}

\subsubsection{How we estimate the convergence rate}
\begin{figure}
\includegraphics[width=\linewidth,height=\textheight,keepaspectratio]{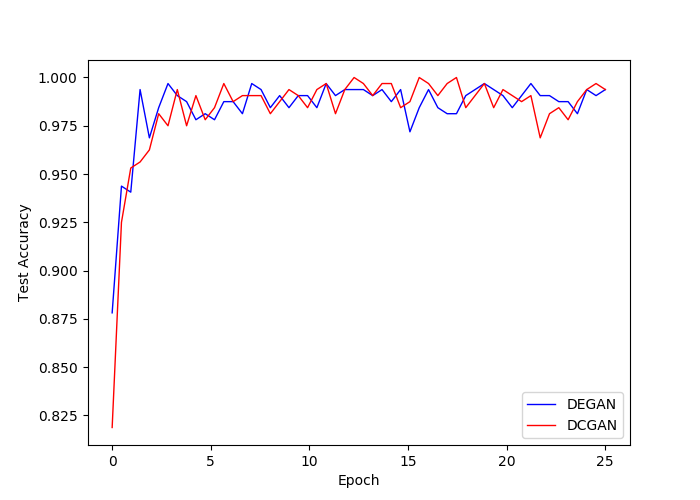}
\caption{As we can see, our DE-GANs model can achieve a test accuracy of 95\% after just one epoch and could reach convergence in only 2 epochs while the DCGANs baseline is slower. The fast convergence rate is resulted from the pre-trained Decoder-Encoder structure which limits the noise latent space. }
\label{figure:convergence}
\end{figure}

As described above, the quality of generated digit images could be evaluated by the test accuracy using a pre-trained classifier. Hence, if we obtain generated images during the training stage, we could use these images to trace image quality changing process. If the quality of images generated by a certain model reaches stability, we can determine the training process has achieved convergence. In our experiments, we sample images each epoch (totally 25 epochs) and use the classifier to evaluate the quality of each iteration. Finally, we find DE-GANs model could reach convergence faster than the baseline DCGANs model. The convergence curves are shown in figure \ref{figure:convergence}

\subsubsection{The diversity of the generated images}
Besides the quality of generated images and the convergence rate of the model, we propose another criterion to measure the capability of generative models, that is the diversity of generated images. As opposed to fixed typewritten digit images, human's written digits are metamorphic from time to time. A single numeral figure could be produced with thousands of patterns. So an important metric on the generalization ability of these generative models is the diversity of the digits generated by them.

Inspired by the concept of variance, we define the diversity degree of a set of digit images as follows:
\begin{definition}
Suppose $X$ is a set of generated image matrices, and $\x_1, \x_2, \cdots, \x_n \in X$ the diversity of the image set is
$$D(X) = \sum_{i=1}^{n}\Vert \x_i-\bar{\x} \Vert^2$$
$$ \bar{\x} = \frac{1}{n}\sum_{i=1}^{n}\x_i $$
\end{definition}
The formula describes the average distance between the generated images to the mean value of them. A larger diversity degree means the model could preserve more types of morphology of handwritten digits. For comparative purposes, we firstly calculate the diversity degree of the handwritten MNIST dataset and then compute that of different generative models. We will use a relative diversity degree(rdd) to compare different models. The rdd is simply a ratio of the diversity degree of a certain model to the MNIST dataset.$$ rdd=\frac{D(X)}{D_{MNIST}} $$ Since there are 10 types of figures in MNIST dataset, we compute the rdd seperately. The results are shown in table \ref{table:diversity}. And we sample a series of generated images from the four models.

\begin{table}
\centering
\begin{tabular}{c c c c c}
\hline
Digit & DCGANs & VAE & WGAN & DE-GANs\\
\hline
0 & 0.952 & 0.836 & 1.007 & 0.995 \\
1 & 0.935 & 0.864 & 2.261 & 0.970 \\
2 & 1.012 & 0.786 & 1.000 & 1.038 \\
3 & 0.923 & 0.778 & 1.150 & 1.023 \\
4 & 1.009 & 0.765 & 1.219 & 1.015 \\
5 & 1.004 & 0.776 & 1.067 & 1.017 \\
6 & 0.987 & 0.812 & 1.167 & 1.113 \\
7 & 0.968 & 0.815 & 1.316 & 0.962 \\
8 & 0.974 & 0.735 & 1.101 & 1.070 \\
9 & 1.002 & 0.788 & 1.310 & 1.011 \\
\hline
Average & 0.977 & 0.796 & 1.260 & 1.021 \\
\hline
\end{tabular}
\caption{The diversity degree of four different generative models, with MNIST original reference 1.0}
\label{table:diversity}
\end{table}
As shown in table \ref{table:diversity}, we can say that our DE-GANs model preserve the inherent structures of the MNIST digit images which performs better than DCGANs and VAE model.

\begin{figure*}
\includegraphics[width=\linewidth,height=\textheight,keepaspectratio]{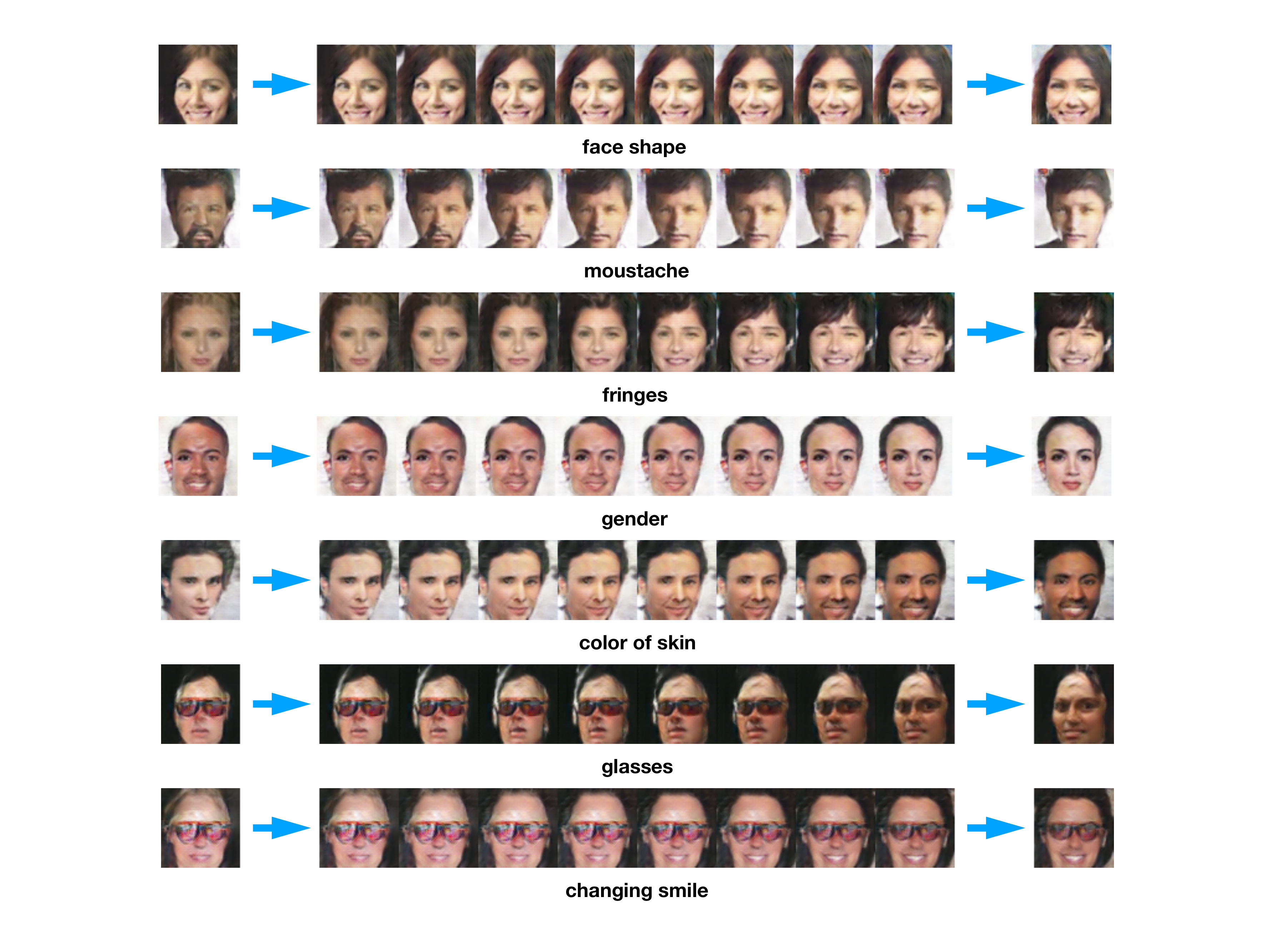}
\caption{Samples from generated images of the DE-GANs model. The facial pictures of every row is generated by sampling a series of noise vectors from the prior distribution. It shows the smooth transition of our learned latent space. The continuously changing input vectors result in the transition of the smooth transition of two different face styles. These changes can be holistic transformation such as the face shape (shown in the first line), or fine adjustment such as the vanishment of moustache (shown in the second line) and the glasses (shown in the 6th line).}
\label{figure:celebA_change_img}
\end{figure*}
\subsection{Experiments on celebA Dataset}
\label{4.2}
\begin{figure}
\includegraphics[width=\linewidth,height=\textheight,keepaspectratio]{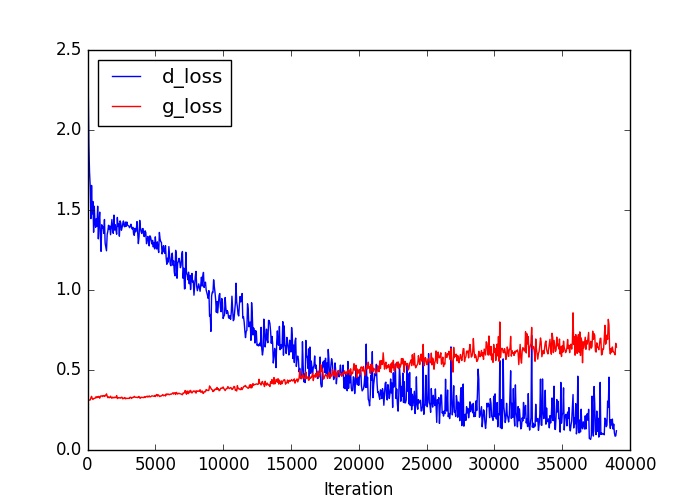}
\label{celebA_DEGAN_line}

\caption{Loss changing figures using celebA dataset as training data.}
\end{figure}
The celebA dataset is a large-scale face image database which contains over 200000 human's facial expressions. It was firstly used for predicting face attributes~\cite{liu2015faceattributes} and face detection~\cite{yang2015facial}. Unlike the MNIST dataset, celebA doesn't consist of explicit labels. Hence, during the training process, we don't add any category information to the input data. In this section, we will mainly view the results of generation and the loss changing during the training process.\\
We collected the loss value data during training and the result is demonstrated in figure \ref{celebA_DEGAN_line} where the $d_{loss}$ phrase means the loss value of the discrimination and the $g_{loss}$ means the loss value of generation. As defined in section \ref{2},
$d_{loss} = \mathfrak{L}_{D}$
and $g_{loss} = \lambda_1 \mathfrak{L}_{adv}+\lambda_2 \mathfrak{L}_{hid}$. \\
The empirical results of DE-GANs model is shown in figure \ref{figure:celebA_change_img}. The figures demonstrate that the DE-GANs model is able to learn the features of the input image dataset and use these attributes to build new images. It builds a bridge between the gap of two seemingly irrelative pictures and makes a reasonable interpolation. This capability can help us generate different face images according to the given input vectors.

\subsection{Experiments on CIFAR-10 Dataset}
\label{4.3}
\begin{figure}
\includegraphics[width=\linewidth,height=\textheight,keepaspectratio]{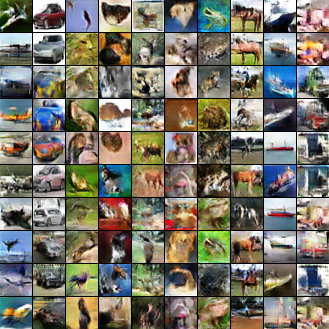}
\caption{These images are samples from the set of generated images using CIFAR-10 as the input real images. The labels from left to right are successively \emph{airplane,automobile,bird,cat,deer,dog,frog,horse,ship,truck.}}
\label{figure:cifar10}
\end{figure}

\begin{figure}

\includegraphics[width=\linewidth,height=\textheight,keepaspectratio]{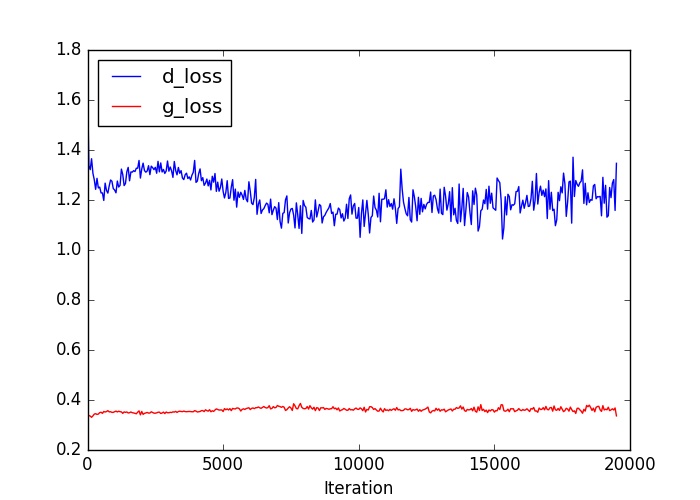}

\caption{The loss value changing curve of DE-GANs}
\label{figure:cifar_change_line}
\end{figure}

In addition to the handwritten digits and human face images, we also explore the DE-GANs model on CIFAR-10 database which contains 60000 $32\times 32$ color images of birds, cat and ship, etc. We use them as the input real data distribution to train the DE-GANs model and the results are presented in figure \ref{figure:cifar10}, which shows the generation ability on tiny images of our DE-GANs model.\\
In order to compare the convergence rate of our DE-GANs model with DCGANs model, we collect $d_{loss}$ and $g_{loss}$ data during the training process, following the analysis on celebA experiments. The results are shown in figure \ref{figure:cifar_change_line}

In order to perceive the quality of these generated images, we trained a classifier~\cite{iandola2014densenet} on CIFAR-10, which can classifies the images to 10 classes, and then used it to measure the quality of the generated images. The DEGANs generated pictures of 10 categories separately and each image is bound with a label. If the pre-trained discriminator could correctly classify an image, we say it is with good quality. The accuracy of classification on the whole datasets reflects the ability of a generative model.

\begin{table}
\centering
\begin{tabular}{c c c c c}
\hline
Class & DCGANs & VAE & DE-GANs\\
\hline
airplane & 0.656 & 0.113 & 0.684  \\
automobile & 0.779 & 0.440 & 0.844  \\
bird & 0.659 & 0.256 & 0.714  \\
cat & 0.602 & 0.460 & 0.648  \\
deer & 0.746 & 0.817 & 0.821  \\
dog & 0.597 & 0.408 & 0.671  \\
frog & 0.874 & 0.282 & 0.869  \\
horse & 0.890 & 0.121 & 0.813  \\
ship & 0.909 & 0.385 & 0.927  \\
truck & 0.807 & 0.565 & 0.809  \\
\hline
Average & 0.752 & 0.385 & 0.780 \\
\hline
\end{tabular}
\caption{The test accuracy of three different models}
\label{table:cifar_acc}
\end{table}
As shown in table \ref{table:cifar_acc}, we compared the generative ability of DEGANs with that of the DCGANs and VAEs model. And a significant test accuracy improvement could be perceived.
\subsection{Visualization on noise}
\begin{figure}
\subfloat[]{
\includegraphics[width=\linewidth,height=\textheight,keepaspectratio]{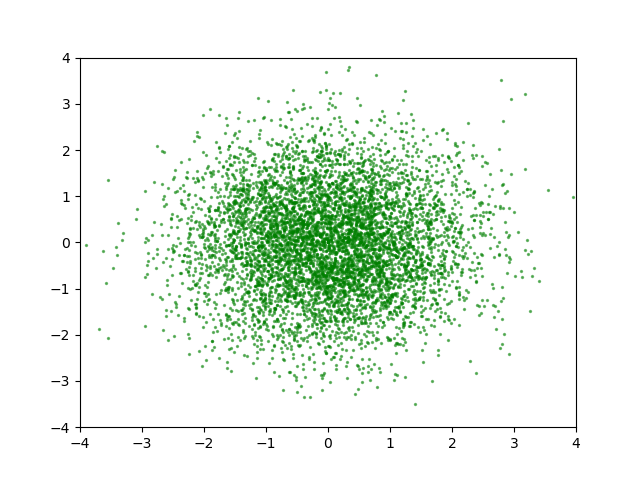}
}
\quad
\subfloat[]{\includegraphics[width=\linewidth,height=\textheight,keepaspectratio]{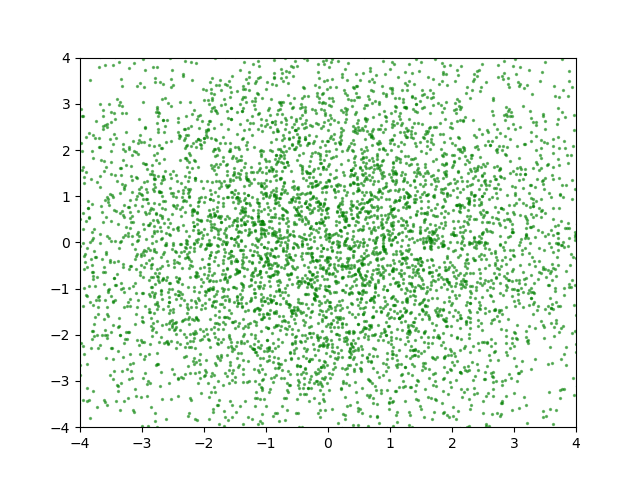}
}

\caption{The top figure shows the embedded points of the original noise vectors while the bottom one is of the posterior noise vectors. As we can see, the original noise basically fits the Gaussian distribution and gathers on the origin of coordinates. But the posterior ones are much more sparse. So we could reach the conclusion that the Decoder-Encoder structure could change the distribution form of the input vectors.}
\label{figure:noise_pca}
\end{figure}

In order to make an exploration of the reason why the Decoder-Encoder structure could improve the quality of the generated images, we sampled the original noise vectors and the corresponding posterior ones during the experiments and use principal component analysis~\cite{dunteman1989principal} method to visualize these noise vectors. They are embedded to a two-dimensional space. The visualization result is presented in figure \ref{figure:noise_pca}.

\subsection{Discussion on the experiments}
As discussed above, the empirical results show the generative ability of the DE-GANs model. \\In \ref{4.1}, we demonstrate that this model can generative relatively high quality of handwritten digit images and 99.1\% of these digits can be recognized by the CNN classifier. We also expound that the Decoder-Encoder structure is able to accelerate the training process of the adversarial networks by adding prior information to the input noise vectors. After only 2 epochs, the classification test accuracy can reach 98\% and its convergence rate is faster than the baseline DCGANs model. We also show that our DE-GANs model could preserve the diversity of original MNIST handwritten digits using the proposed \emph{rdd} concept, which is a metric to make a seperation of handwritten characters and typewritten ones. In this term, DE-GANs outperforms the state-of-art model DCGANs. \\ In \ref{4.2}, we present the generated images of human face and it shows an impressive result of the transform between two seemingly irrelative faces. We can see the learned latent space of DE-GANs generator has a smooth space structure. \\ In \ref{4.3}, we see the generated results on tiny natural color images and it shows the capability of learning these kinds of data. The quality of the generated images is significantly higher than that of the baseline DCGANs model.

\section{Conclusion}
\label{5}
In this work, we proposed a new generative model DE-GANs, which show an outstanding performance on image generation. Before the input noise vectors passed into the generator of the adversarial networks, it flows through a pre-trained Decoder-Encoder structure using the original image dataset. This structure is pre-trained using the Bayesian variational inference, so the latent space of the structure contains information of the image dataset. Hence, the input vectors of the generator are limited to a prior distribution which is more compact than the Gaussian distribution. \\
In section \ref{4}, the empirical results show that these prior input vectors can help accelerate the training process of the DE-GANs model and preserve the diversity of the original MNIST handwritten figures. In this terms, the DE-GANs model outperforms the state-of-the-art DCGANs model. Besides the MNIST dataset, experiments on celebA and CIFAR-10 also show the learned space of the DE-GANs generator has a smooth inherent structure. The generated results of human face images indicate an impressive transformation between two different style of facial expressions. We also used a combined loss function to improve the stability of generation and accelerate the training process.


%





\ifCLASSOPTIONcaptionsoff
  \newpage
\fi



%

\bibliographystyle{IEEEtranS}
\bibliography{ref}

\begin{IEEEbiographynophoto}{Guoqiang Zhong}
received his BS degree in Mathematics from Hebei Normal University, Shijiazhuang, China, his MS degree in Operations Research and Cybernetics from Beijing University of Technology (BJUT), Beijing, China, and his PhD in Pattern Recognition and Intelligent Systems from Institute of Automation, Chinese Academy of Sciences (CASIA), Beijing, China, in 2004, 2007 and 2011, respectively. Between October 2011 and July 2013, he was a Postdoctoral Fellow with the Synchromedia Laboratory for Multimedia Communication in Telepresence, \'{E}cole de Technologie Sup\'{e}rieure (ETS), University of Quebec, Montreal, Canada. Since March 2014, he has been an associate professor at Department of Computer Science and Technology, Ocean University of China, Qingdao, China. Dr. Zhong has published more than 50 technical papers in the areas of artificial intelligence, pattern recognition, machine learning and data mining. His research interests include pattern recognition, machine learning and image processing.
\end{IEEEbiographynophoto}


\end{document}

%% file: packages.tex
\def\I{{\bf I}}

\def\x{{\bf x}}

\def\z{{\bf z}}

\def\0{{\bf 0}}
\def\1{{\bf 1}}

%% file: main.bbl
\begin{thebibliography}{10}
\providecommand{\url}[1]{#1}
\csname url@samestyle\endcsname
\providecommand{\newblock}{\relax}
\providecommand{\bibinfo}[2]{#2}
\providecommand{\BIBentrySTDinterwordspacing}{\spaceskip=0pt\relax}
\providecommand{\BIBentryALTinterwordstretchfactor}{4}
\providecommand{\BIBentryALTinterwordspacing}{\spaceskip=\fontdimen2\font plus
\BIBentryALTinterwordstretchfactor\fontdimen3\font minus
  \fontdimen4\font\relax}
\providecommand{\BIBforeignlanguage}[2]{{%
\expandafter\ifx\csname l@#1\endcsname\relax
\typeout{** WARNING: IEEEtranS.bst: No hyphenation pattern has been}%
\typeout{** loaded for the language `#1'. Using the pattern for}%
\typeout{** the default language instead.}%
\else
\language=\csname l@#1\endcsname
\fi
#2}}
\providecommand{\BIBdecl}{\relax}
\BIBdecl

\bibitem{arjovsky2017wasserstein}
M.~Arjovsky, S.~Chintala, and L.~Bottou, ``Wasserstein gan,'' \emph{arXiv
  preprint arXiv:1701.07875}, 2017.

\bibitem{bengio2013representation}
Y.~Bengio, A.~Courville, and P.~Vincent, ``Representation learning: A review
  and new perspectives,'' \emph{IEEE transactions on pattern analysis and
  machine intelligence}, vol.~35, no.~8, pp. 1798--1828, 2013.

\bibitem{bennett1969intrinsic}
R.~Bennett, ``The intrinsic dimensionality of signal collections,'' \emph{IEEE
  Transactions on Information Theory}, vol.~15, no.~5, pp. 517--525, 1969.

\bibitem{denton2015deep}
E.~L. Denton, S.~Chintala, R.~Fergus \emph{et~al.}, ``Deep generative image
  models using a laplacian pyramid of adversarial networks,'' in \emph{Advances
  in neural information processing systems}, 2015, pp. 1486--1494.

\bibitem{dorigo2006ant}
M.~Dorigo, M.~Birattari, and T.~Stutzle, ``Ant colony optimization,''
  \emph{IEEE computational intelligence magazine}, vol.~1, no.~4, pp. 28--39,
  2006.

\bibitem{dosovitskiy2017learning}
A.~Dosovitskiy, J.~T. Springenberg, M.~Tatarchenko, and T.~Brox, ``Learning to
  generate chairs, tables and cars with convolutional networks,'' \emph{IEEE
  transactions on pattern analysis and machine intelligence}, vol.~39, no.~4,
  pp. 692--705, 2017.

\bibitem{dosovitskiy2015learning}
A.~Dosovitskiy, J.~Tobias~Springenberg, and T.~Brox, ``Learning to generate
  chairs with convolutional neural networks,'' in \emph{Proceedings of the IEEE
  Conference on Computer Vision and Pattern Recognition}, 2015, pp. 1538--1546.

\bibitem{du2017learning}
C.~Du, J.~Zhu, and B.~Zhang, ``Learning deep generative models with doubly
  stochastic gradient mcmc,'' \emph{IEEE transactions on neural networks and
  learning systems}, 2017.

\bibitem{dunteman1989principal}
G.~H. Dunteman, \emph{Principal components analysis}.\hskip 1em plus 0.5em
  minus 0.4em\relax Sage, 1989, no.~69.

\bibitem{efros1999texture}
A.~A. Efros and T.~K. Leung, ``Texture synthesis by non-parametric sampling,''
  in \emph{Computer Vision, 1999. The Proceedings of the Seventh IEEE
  International Conference on}, vol.~2.\hskip 1em plus 0.5em minus 0.4em\relax
  IEEE, 1999, pp. 1033--1038.

\bibitem{freeman2002example}
W.~T. Freeman, T.~R. Jones, and E.~C. Pasztor, ``Example-based
  super-resolution,'' \emph{IEEE Computer graphics and Applications}, vol.~22,
  no.~2, pp. 56--65, 2002.

\bibitem{goh2014learning}
H.~Goh, N.~Thome, M.~Cord, and J.-H. Lim, ``Learning deep hierarchical visual
  feature coding,'' \emph{IEEE transactions on neural networks and learning
  systems}, vol.~25, no.~12, pp. 2212--2225, 2014.

\bibitem{goodfellow2014generative}
I.~Goodfellow, J.~Pouget-Abadie, M.~Mirza, B.~Xu, D.~Warde-Farley, S.~Ozair,
  A.~Courville, and Y.~Bengio, ``Generative adversarial nets,'' in
  \emph{Advances in neural information processing systems}, 2014, pp.
  2672--2680.

\bibitem{gregor2015draw}
K.~Gregor, I.~Danihelka, A.~Graves, D.~J. Rezende, and D.~Wierstra, ``Draw: A
  recurrent neural network for image generation,'' \emph{arXiv preprint
  arXiv:1502.04623}, 2015.

\bibitem{gurumurthy2017deligan}
S.~Gurumurthy, R.~K. Sarvadevabhatla, and V.~B. Radhakrishnan, ``Deligan:
  Generative adversarial networks for diverse and limited data,'' in \emph{The
  IEEE Conference on Computer Vision and Pattern Recognition (CVPR)}, vol.~1,
  2017.

\bibitem{hays2007scene}
J.~Hays and A.~A. Efros, ``Scene completion using millions of photographs,'' in
  \emph{ACM Transactions on Graphics (TOG)}, vol.~26, no.~3.\hskip 1em plus
  0.5em minus 0.4em\relax ACM, 2007, p.~4.

\bibitem{huang2017beyond}
R.~Huang, S.~Zhang, T.~Li, and R.~He, ``Beyond face rotation: Global and local
  perception gan for photorealistic and identity preserving frontal view
  synthesis,'' \emph{arXiv preprint arXiv:1704.04086}, 2017.

\bibitem{iandola2014densenet}
F.~Iandola, M.~Moskewicz, S.~Karayev, R.~Girshick, T.~Darrell, and K.~Keutzer,
  ``Densenet: Implementing efficient convnet descriptor pyramids,'' \emph{arXiv
  preprint arXiv:1404.1869}, 2014.

\bibitem{ioffe2015batch}
S.~Ioffe and C.~Szegedy, ``Batch normalization: Accelerating deep network
  training by reducing internal covariate shift,'' in \emph{International
  Conference on Machine Learning}, 2015, pp. 448--456.

\bibitem{kingma2013auto}
D.~P. Kingma and M.~Welling, ``Auto-encoding variational bayes,'' \emph{arXiv
  preprint arXiv:1312.6114}, 2013.

\bibitem{krizhevsky2009learning}
A.~Krizhevsky and G.~Hinton, ``Learning multiple layers of features from tiny
  images,'' University of Toronto, Technical report, 2009.

\bibitem{krizhevsky2012imagenet}
A.~Krizhevsky, I.~Sutskever, and G.~E. Hinton, ``Imagenet classification with
  deep convolutional neural networks,'' in \emph{Advances in neural information
  processing systems}, 2012, pp. 1097--1105.

\bibitem{kullback1951information}
S.~Kullback and R.~A. Leibler, ``On information and sufficiency,'' \emph{The
  annals of mathematical statistics}, vol.~22, no.~1, pp. 79--86, 1951.

\bibitem{larsen2015autoencoding}
A.~B.~L. Larsen, S.~K. S{\o}nderby, H.~Larochelle, and O.~Winther,
  ``Autoencoding beyond pixels using a learned similarity metric,'' \emph{arXiv
  preprint arXiv:1512.09300}, 2015.

\bibitem{MNIST}
Y.~Lecun, ``The mnist database of handwritten digits,'' 1998, data retrieved
  from \url{http://yann.lecun.com/exdb/mnist/}.

\bibitem{levina2005maximum}
E.~Levina and P.~J. Bickel, ``Maximum likelihood estimation of intrinsic
  dimension,'' in \emph{Advances in neural information processing systems},
  2005, pp. 777--784.

\bibitem{liu2015faceattributes}
Z.~Liu, P.~Luo, X.~Wang, and X.~Tang, ``Deep learning face attributes in the
  wild,'' in \emph{Proceedings of International Conference on Computer Vision
  (ICCV)}, Dec. 2015.

\bibitem{makhzani2017pixelgan}
A.~Makhzani and B.~J. Frey, ``Pixelgan autoencoders,'' in \emph{Advances in
  Neural Information Processing Systems}, 2017, pp. 1972--1982.

\bibitem{makhzani2015adversarial}
A.~Makhzani, J.~Shlens, N.~Jaitly, I.~Goodfellow, and B.~Frey, ``Adversarial
  autoencoders,'' \emph{arXiv preprint arXiv:1511.05644}, 2015.

\bibitem{masci2011stacked}
J.~Masci, U.~Meier, D.~Cire{\c{s}}an, and J.~Schmidhuber, ``Stacked
  convolutional auto-encoders for hierarchical feature extraction,'' in
  \emph{International Conference on Artificial Neural Networks}.\hskip 1em plus
  0.5em minus 0.4em\relax Springer, 2011, pp. 52--59.

\bibitem{mescheder2017adversarial}
L.~Mescheder, S.~Nowozin, and A.~Geiger, ``Adversarial variational bayes:
  Unifying variational autoencoders and generative adversarial networks,''
  \emph{arXiv preprint arXiv:1701.04722}, 2017.

\bibitem{portilla2000parametric}
J.~Portilla and E.~P. Simoncelli, ``A parametric texture model based on joint
  statistics of complex wavelet coefficients,'' \emph{International journal of
  computer vision}, vol.~40, no.~1, pp. 49--70, 2000.

\bibitem{radford2015unsupervised}
A.~Radford, L.~Metz, and S.~Chintala, ``Unsupervised representation learning
  with deep convolutional generative adversarial networks,'' \emph{arXiv
  preprint arXiv:1511.06434}, 2015.

\bibitem{setiono2002extraction}
R.~Setiono, W.~K. Leow, and J.~M. Zurada, ``Extraction of rules from artificial
  neural networks for nonlinear regression,'' \emph{IEEE transactions on neural
  networks}, vol.~13, no.~3, pp. 564--577, 2002.

\bibitem{sohl2015deep}
J.~Sohl-Dickstein, E.~A. Weiss, N.~Maheswaranathan, and S.~Ganguli, ``Deep
  unsupervised learning using nonequilibrium thermodynamics,'' \emph{arXiv
  preprint arXiv:1503.03585}, 2015.

\bibitem{van2016conditional}
A.~van~den Oord, N.~Kalchbrenner, L.~Espeholt, O.~Vinyals, A.~Graves
  \emph{et~al.}, ``Conditional image generation with pixelcnn decoders,'' in
  \emph{Advances in Neural Information Processing Systems}, 2016, pp.
  4790--4798.

\bibitem{wei2002multisynapse}
C.-H. Wei and C.-S. Fahn, ``The multisynapse neural network and its application
  to fuzzy clustering,'' \emph{IEEE transactions on neural networks}, vol.~13,
  no.~3, pp. 600--618, 2002.

\bibitem{xing2017deep}
F.~Xing, Y.~Xie, H.~Su, F.~Liu, and L.~Yang, ``Deep learning in microscopy
  image analysis: A survey,'' \emph{IEEE Transactions on Neural Networks and
  Learning Systems}, 2017.

\bibitem{yang2015facial}
S.~Yang, P.~Luo, C.-C. Loy, and X.~Tang, ``From facial parts responses to face
  detection: A deep learning approach,'' in \emph{Proceedings of the IEEE
  International Conference on Computer Vision}, 2015, pp. 3676--3684.

\bibitem{yu2009nonlinear}
K.~Yu, T.~Zhang, and Y.~Gong, ``Nonlinear learning using local coordinate
  coding,'' in \emph{Advances in neural information processing systems}, 2009,
  pp. 2223--2231.

\bibitem{zeiler2014visualizing}
M.~D. Zeiler and R.~Fergus, ``Visualizing and understanding convolutional
  networks,'' in \emph{European conference on computer vision}.\hskip 1em plus
  0.5em minus 0.4em\relax Springer, 2014, pp. 818--833.

\bibitem{zeiler2010deconvolutional}
M.~D. Zeiler, D.~Krishnan, G.~W. Taylor, and R.~Fergus, ``Deconvolutional
  networks,'' in \emph{Computer Vision and Pattern Recognition (CVPR), 2010
  IEEE Conference on}.\hskip 1em plus 0.5em minus 0.4em\relax IEEE, 2010, pp.
  2528--2535.

\end{thebibliography}
